\documentclass[letterpaper]{article} 
\usepackage{aaai19}  
\usepackage{times}  
\usepackage{helvet}  
\usepackage{courier}  
\usepackage{url}  
\usepackage{graphicx}  
\author{Keigo Kawamura \and Yoshimasa Tsuruoka \\
The University of Tokyo \\
\{kkawamura, tsuruoka\}@logos.t.u-tokyo.ac.jp
}
\title{Neural Fictitious Self-Play on ELF Mini-RTS}

\usepackage{amsmath}
\usepackage{amsfonts}
\usepackage{bm}
\usepackage{algorithm, algpseudocode}

\usepackage{subcaption}

\newcommand{\citet}[1]
{\citeauthor{#1} ̃\shortcite{#1}}
\newcommand{\citep}{\cite}

\newcommand{\middlebar}{\,|\,}

\makeatletter
\newcommand*{\rom}[1]{\expandafter\@slowromancap\romannumeral #1@}
\makeatother

\begin{document}
\maketitle
\begin{abstract}
  Despite the notable successes in video games such as Atari 2600, current AI is yet to defeat human champions in the domain of real-time strategy (RTS) games. One of the reasons is that an RTS game is a multi-agent game, in which single-agent reinforcement learning methods cannot simply be applied because the environment is not a stationary Markov Decision Process. In this paper, we present a first step toward finding a game-theoretic solution to RTS games by applying Neural Fictitious Self-Play (NFSP), a game-theoretic approach for finding Nash equilibria, to Mini-RTS, a small but nontrivial RTS game provided on the ELF platform.
  More specifically, we show that NFSP can be effectively combined with policy gradient reinforcement learning and be applied to Mini-RTS. Experimental results also show that the scalability of NFSP can be substantially improved by pretraining the models with simple self-play using policy gradients, which by itself gives a strong strategy despite its lack of theoretical guarantee of convergence.
\end{abstract}

\section{Introduction}

\noindent With the recent rise of deep neural networks, reinforcement learning has shown remarkable achievements in many complex environments.
In the Atari 2600 video game environment, agents trained with deep reinforcement learning methods have succeeded in achieving human-level, or even super-human performance in most of the games~\citep{DQN2015,a3c2016,HybridRewardArchitecture2017}.
However, in the domain of real-time strategy (RTS) games, which are considered to be one of the next grand AI challenges after Chess and Go~\citep{elf2017,AlphaZero2017}, current AI is yet to defeat top human players~\citep{StarCraft2_2017}.

To tackle this challenging domain, several platforms for conducting experiments on RTS games have been developed~\citep{muRTS2013,TorchCraft2016,StarCraft2_2017}. The ELF platform~\citep{elf2017} is such a platform and is an extensive, lightweight, and flexible platform designed for reinforcement learning research. It provides a small but nontrivial RTS game called Mini-RTS, and this game runs an order of magnitude faster than existing RTS environments, while capturing all the basic dynamics of RTS games, e.g., fog-of-war, resource gathering, troop building, and attacking with troops.

In this work, we aim to find a game-theoretic solution to Mini-RTS; that is, we attempt to compute an equilibrium strategy profile, as a first step toward solving more realistic and complex RTS games. Developing an AI for RTS involves many difficulties, including strategic and tactical decision making, real-time planning, and domain knowledge exploitation~\citep{RTSSurvey2013,RTSSurvey2014}. In this paper, we particularly focus on the multi-agent property: RTS games are multi-agent games and thus are not stationary for a learning agent, which breaks an assumption of single-agent reinforcement learning that the environment can be modeled as a Markov Decision Process (MDP).

\citet{FSP2015} proposed a game-theoretic self-play approach called Fictitious Self-Play (FSP). In FSP, an agent calculates the best response strategy to its opponents with reinforcement learning and averages its strategies in a sampling-based fashion. This process forms Fictitious Play (FP)~\citep{FP1951,GWFP2006} in extensive-form games, and can be applied to a large-scale imperfect-information game.
Since FP has a theoretical guarantee of convergence to a Nash equilibrium with minimal restrictions, FSP has a reliable theoretical background on convergence, and is more likely to converge than raw self-play methods.
Neural Fictitious Self-Play (NFSP)~\citep{NFSP2016}, a variant of FSP that uses deep reinforcement learning for its best response component, learned an approximate Nash equilibrium in small games of Poker without any prior domain knowledge.

In this paper, we show that NFSP can be effectively combined with policy gradient reinforcement learning and be used in the Mini-RTS domain. Our experimental results also show that the scalability of NFSP can be substantially improved by pretraining the models with simple self-play using a policy gradient method, which is efficient and by itself gives a strong strategy despite its lack of theoretical guarantee of convergence.
To the best of our knowledge, this is the first attempt to find a convergent strategy profile in a non-trivial RTS game, hereby presenting a promising direction toward finding Nash equilibrium strategies for RTS games (i.e., solving RTS games).

\section{Task}

\subsection{ELF and Mini-RTS}

Our objective is to compute an equilibrium strategy that is not exploitable for the Mini-RTS game in the ELF platform.

In Mini-RTS, the goal of the agent is to destroy the opponent's base with its troops. Each agent has its base, units, and resource. With its base and some resource, the agent can build a worker. A worker can build a barrack, and some attackers with the barrack.

\begin{figure}[t]
  \centering
  \includegraphics[width=40mm]{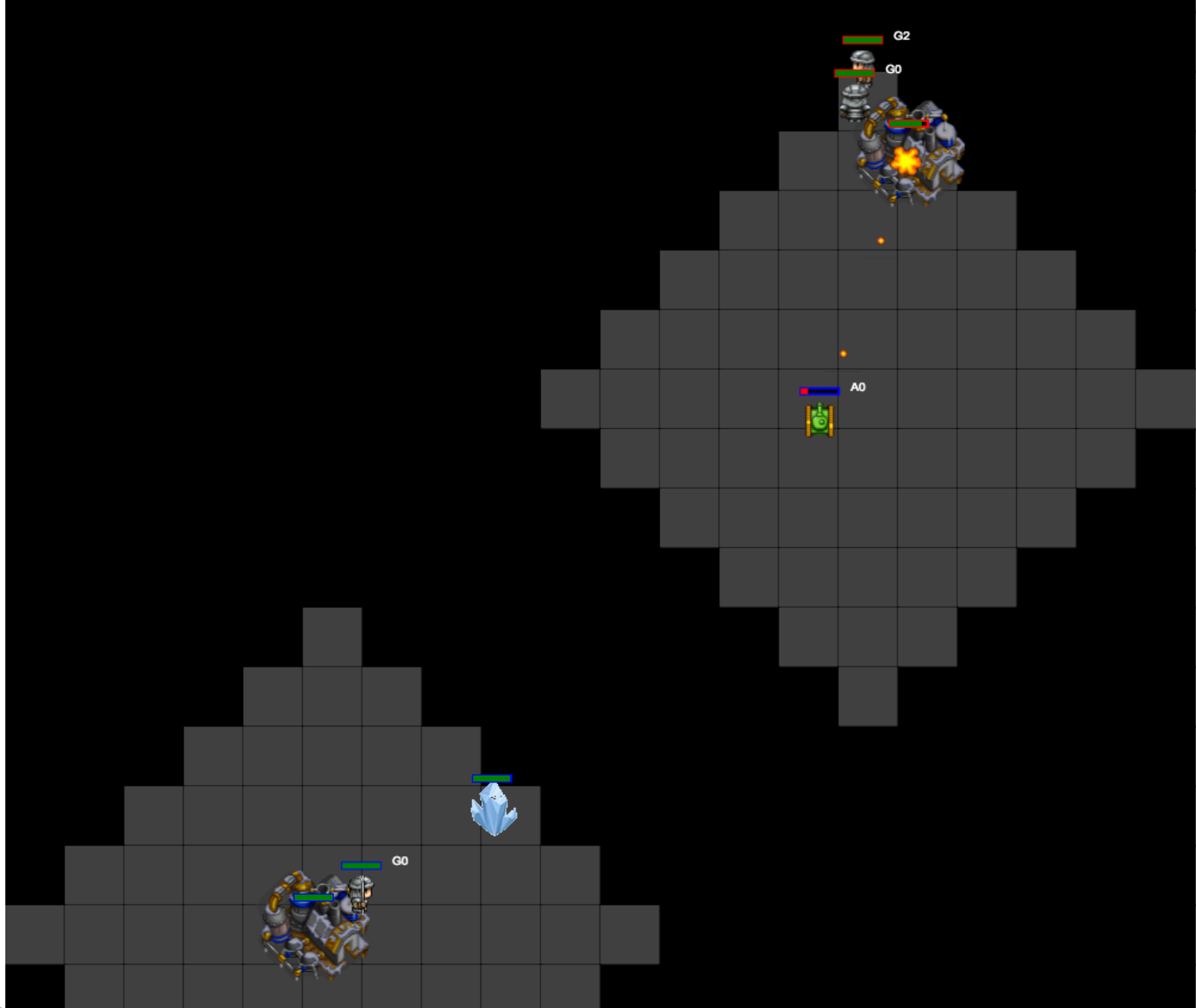}
  \caption{A game screenshot of Mini-RTS. Here the sight is for the lower-left agent whose health bars are enclosed by a blue line. Because of the fog-of-war, the agent cannot see the vicnity of the opponent's base, so it does not know whether the opponent has any troops or not.}
  \label{fig:example_mini_rts}
\end{figure}

The ELF game engine is tick-driven: at each tick, each agent makes decisions by sending commands on units based on the observation. The game state changes according to the commands and new observations are given to the agents. Because there is fog-of-war in mini-RTS as in other RTS games, agents cannot observe units of its opponents in fog-of-war, and thus the game is imperfect information.
A screenshot of the game is shown in Figure~\ref{fig:example_mini_rts}.

In addition to the low- and micro-level commands like ``move left by two pixels'' for each unit, the ELF engine has more hierarchical and strategic commands like ``make someone go to some available place and build a barrack'' or ``defend our base from the enemy's attackers'' for its all units. We use these commands instead of raw commands.
Specifically, agents have nine discrete strategic actions. Four of these are about building units: a worker, a barrack, a melee attacker, and a range attacker. The next four are about tactical commands: attack, attack-in-range, hit-and-run, and all-defend. The last command is Idle, which means doing nothing. These actions are global, i.e., they affect all units in one command.
Agents receive a low-level observation matrix shaped $22\times 20 \times 20$ at each tick, where $20\times 20$ represents the resolution of the observation for the game map, and $22$ channels contain the number of each kind of units like a worker or the base.

\section{Background}

\subsection{Markov Decision Process and Reinforcement Learning}

An MDP is an environment model for standard reinforcement learning.
In reinforcement learning with an MDP, an agent interacts with an MDP environment $\mathcal{E}$. At each time step $t$, the agent receives a state $s_t \in \mathcal{S}$ and selects an action $a_t$ from a set of possible actions $\mathcal{A}$ with a probability distribution $\pi: \mathcal{S} \times \mathcal{A} \rightarrow \left[0, 1\right]$, which is called a policy. In $\mathcal{E}$ the action is executed, and it returns a next state $s_{t+1}$ with reward $r_{t+1}$. The goal of the agent is to maximize its expected cumulative reward $\mathbb{E}\left[R_t\right] = \mathbb{E} \left[\sum_{k=0}^\infty \gamma^k r_{t+k}\right]$, where $\gamma$ is a discount factor.

If the agent cannot distinguish between some states of the environment, the environment is called a Partially Observable MDP (POMDP). In a POMDP environment $\mathcal{E}_p$, the agent receives an observation $o_t=O(s_t)$, where $s_t$ is a true state of $\mathcal{E}_p$ and $O$ is a function that maps a state to an observation of the agent. The agent selects an action $a_t$ from $\mathcal{A}$ as in an MDP, but the policy $\pi$ depends on the observation $o_t$ and not on the state $s_t$, because the agent cannot observe the true state.

We will use the following standard definitions of the state-action value function $Q_\pi \left(s_t, a_t\right) = \mathbb{E}\left[\sum_{k=0}^\infty \gamma^k r_{t+k}\right]$, the value function $V_\pi \left(s_t\right) = \sum_{a\in \mathcal{A}} \pi\left(a\middlebar s_t\right) Q_\pi\left(s_t, a\right)$, and the advantage function $A_\pi \left(s_t, a_t\right) = Q_\pi \left(s_t, a_t\right) - V_\pi \left(s_t\right)$.

\subsection{Extensive-Form Games}

In this work, we regard RTS games as extensive-form games. An extensive-form game is a model for a sequential multi-agent game. The representation is based on a finite rooted game tree.

In an extensive-form game, for each agent $i\in \mathcal{N}$, there are some indistinguishable states. An information set $u_i \in U_i$ contains such states; namely, the agent $i$ cannot distinguish $s_1$ and $s_2$ and is forced to act in the exact same way if the two states are in the same information set $u_i$.
If the agents in the game never forget their acquired information, the game is called perfect recall. In a perfect recall game, the graph of information sets forms a tree. And if the game has only two agents and $R_1 + R_2 = 0$ for all states, the game is called a two-player zero-sum game.

Each agent has its own strategy $\pi_i$, which specifies the probability distribution over the possible actions $A(s)$ in the given state $s$. A strategy profile $\pi = \left\{\pi_1, \cdots, \pi_N\right\}$ is a tuple of strategies for all agents. We can consider the expected cumulative reward $R(\pi)$ given a fixed strategy profile. A strategy of the agent $i$ is called the best response strategy to its opponents' strategy $\pi_{-i}=\left\{\pi_j \middlebar j\neq i\right\}$ if the strategy maximizes its expected reward $R\left(\cdot, \pi_{-i}\right)$. A Nash equilibrium of an extensive-form game is a strategy profile such that for each agent its strategy is the best response strategy to the others' strategies.

We can combine this game-theoretic model with MDPs. In an extensive-form game,  if we pick an agent and fix other agents' strategies, then the environment can be regarded as a single-agent POMDP for the picked agent. In addition, if the game is perfect recall, this POMDP can be converted into an MDP environment, because the probability distribution of reaching indistinguishable states is stationary and thus these states can be degenerated into one state.

\subsection{Neural Fictitious Self-Play}

NFSP~\cite{NFSP2016} is a variant of FSP that uses neural networks and Deep-Q Networks (DQN)~\citep{DQN2015} for its approximation functions. FSP is a scalable method that uses FP in extensive-form representations.

In FP, a popular game-theoretic model of learning, agents repeatedly play a game, choosing the best response strategy to their opponents' average strategies at each iteration. The average strategies converge to a Nash equilibrium when the game has certain properties, e.g., two-player zero-sum or potential games.

FP is a theory on a normal-form representation, where each agent acts only once per one game, which is not suited to large-scale applications. To overcome the limitation, \citet{FSP2015} proposed a full-width extensive-form fictitious play and FSP. Both methods are developed for an extensive-form representation, and the former is a full-width method and the latter is an appropriately approximated (hence scalable to large-scale games) method.
As with FP, agents in FSP repeatedly play a game, storing their experience in memory. Instead of computing the full-width best response strategy, they learn an approximate best response using reinforcement learning (RL). And instead of averaging their full-width strategies, they learn an approximate average strategy by using supervised learning (SL).

NFSP is not a method that simply applied neural networks to FSP. In NFSP, agents memorize their experiences in a reservoir replay buffer~\citep{Reservoir1985} to avoid windowing experiences due to sampling from a finite memory. NFSP also uses anticipatory dynamics~\citep{DFSP2005} to enable each agent to effectively track changes in its opponents' behavior.

The resulting NFSP algorithm is as follows. Each agent $i$ has its RL network $\beta_i$, SL network $\pi_i$, and SL reservoir replay buffer $\mathcal{M}_i^{SL}$. At the beginning of the game, each agent decides whether it uses $\beta_i$ or $\pi_i$ as its strategy in this episode, with probability $\eta$ and $1-\eta$, respectively. At each time step, agents sample an action from the selected strategy, and if the selected strategy is $\beta_i$, a tuple of the observation and the taken action is stored in $\mathcal{M}_i^{SL}$. $\beta_i$ is trained as it maximizes the expected cumulative reward against $\pi_{-i}$ and $\pi_i$ is trained as it represents the probability distribution over actions in $\mathcal{M}_i^{SL}$.
Since $\mathcal{M}_i^{SL}$ is a reservoir buffer and tuples in $\mathcal{M}_i^{SL}$ are taken from $\beta_i$, $\pi_i$ demonstrates the average strategy over the past RL strategies.
In addition, since the mixed strategy representated by choosing one from two extensive-form strategies at the beginning of episodes forms a realization equivalent strategy to the mixed strategy of the two normal-form strategy, the behavior strategy in each episode is $\eta \beta^t_i + (1-\eta)\pi^t_i = \pi^t_i + \eta (\beta^t_i - \pi^t_i) \simeq \pi^t_i + \eta \alpha \frac{d}{dt} \pi^t_i \simeq \pi_i^{t+\Delta t} $, which is a short-term prediction of $\pi^t_i$. Thus, for any agent $i$, it computes an approximated best response strategy to its opponents' average strategy (with some time prediction) $\pi_{-i}$, and an approximated average strategy over the past best response strategies $\beta_i$, which forms an approximated FP.

\subsection{Proximal Policy Optimization}

For the reason discussed later in the Method section, we do not use a value-based reinforcement learning method such as DQN as our reinforcement learning algorithm. Instead, we use a policy-based reinforcement learning method called Proximal Policy Optimization (PPO)~\citep{PPO2017}, which extends and simplifies Trust Region Policy Optimization (TRPO)~\citep{TRPO2015}.

In TRPO, an objective function $\mathbb{E}_t\left[\frac{\pi_\theta\left(a_t\middlebar s_t\right)}{\pi_{\theta_{\text{old}}}\left(a_t\middlebar s_t\right)}\hat{A}_t\right]$ is maximized, subject to a constraint on the policy update represented as the Kullback-Leibler divergence between $\pi_\theta\left(\cdot \middlebar s_t\right)$ and $\pi_{\theta_{\text{old}}}\left(\cdot \middlebar s_t\right)$, where $\theta_{\text{old}}$ is the fixed parameters before the update.

The constraint in this optimization problem is introduced to prevent an excessively large policy update. PPO uses a clipping term instead of this constraint, i.e., maximizing the following function under the unconstrained condition:

\begin{align}
  \label{eq:PPO_loss}
  L(\theta) = \hat{\mathbb{E}}_t \left[\min \left(r_t \hat{A}_t, \, \mathop{\rm clip}\left(r_t, \, 1-\epsilon, \, 1+\epsilon\right)\hat{A}_t\right)\right]\, ,
\end{align}
where $r_t$ is the probability ratio $r_t = \frac{\pi_\theta \left(a_t\middlebar s_t\right)}{\pi_{\theta_{\text{old}}}\left(a_t\middlebar s_t\right)}$, $\epsilon$ is a hyperparameter that determines the threshold, and $\hat{A}_t$ is an estimator of the advantage function $A_t$. This scheme is much simpler to implement and empirically has better performance than original TRPO.

\section{Method}

We regard RTS games as two-player zero-sum perfect recall extensive-form games and apply self-play methods to them. This view is justified as follows:
although there are many units in an RTS game, agents can control all of them and have all information about them, and hence the game is essentially a two-player zero-sum game.
Because of the existence of fog-of-war, the game is an imperfect information game. An RTS game is originally real-time and is not tick-driven, but in practice almost all of the RTS games have discrete time steps and therefore the game can be regarded as an extensive-form game.
If each agent collects all observation histories and treats the set of them as a new observation, then the game is modeled as a perfect recall game. In this work we do not memorize past histories. We will discuss it in the section of future work.

\begin{algorithm}[t!]
  \caption{NFSP with PPO}
  \label{alg:NFSP_PPO}
  \begin{algorithmic}[1]
    \Statex $\Gamma$ is an ELF interface and $N$ is the number of agents
    \Function{Main}{$\Gamma$, $N$}
    \For{$p = 1, 2, \cdots, N$ in parallel} \Comment{$p$ is a learning agent}
      \State $\Gamma$.Initialize()
      \State $\Gamma$.RegisterCallback($p$, Trainer)
      \State $\Gamma$.RegisterCallback($p$, RL\_Actor)
      \For{$q = 1, \cdots, p-1, p+1, \cdots, N$ }
        \State $\Gamma$.RegisterCallback($q$, SL\_Actor)
      \EndFor
      \Repeat
        \Repeat \label{line:do_step_begin}
        \State batch$\gets\Gamma$.StepAndAccumulate() \Comment{Multiple games are executed asyncronously and observation data is accumulated into the batch}
        \Until{The number of accumulated data reaches certain batch size}
        \State $\Gamma$.CorrespondingCallback(batch) \label{line:do_step_end}
      \Until{Time steps exceed the certain limits}
    \EndFor
    \EndFunction

    \Function{Trainer}{$p$, batch}
      \State $\left\{S_\tau, A_\tau, \Pi_\tau, R_\tau\right\}_{\left\{\tau=t,\cdots,t+T-1\right\}} \gets$ batch
      \label{line:batch_trainer}

      \Comment{State, action, probability distribution, and reward}

      \Comment{$\Pi_t$ is a probability distribution of $\text{NN}_{RL}$ at $t$}
      \State Calculate $\mathcal{L}_{RL}$ with eq. (\ref{eq:rl_cost})
      \label{line:calc_rl_loss}
      \State Memorize $\left\{S_{\tau}, \Pi_{\tau}\right\}$ in buffer $\mathcal{M}_{SL}$
      \label{line:memorize_exp}
      \State Sample $S, \Pi \gets \mathcal{M}_{SL}$
      \label{line:get_batch_from_sl_memory}
      \State Calculate $\mathcal{L}_{SL}$ with eq. (\ref{eq:sl_cost})
      \label{line:calc_sl_loss}
      \State Optimize $\text{NN}_{RL}$ and $\text{NN}_{SL}$ with $\mathcal{L}_{RL}$ and $\mathcal{L}_{SL}$
    \EndFunction
    \Function{RL\_Actor}{$p$, batch}
      \State $S \gets$ batch
      \label{line:batch_RL}
      \State $\pi \gets$ $\text{NN}_{RL}(S)$
      \State \Return Sampled $a \gets \pi$
    \EndFunction
    \Function{SL\_Actor}{$p$, batch}
      \State $S \gets$ batch
      \label{line:batch_SL}
      \State $\pi \gets$ $\text{NN}_{SL}(S)$
      \State \Return Sampled $a \gets \pi$
    \EndFunction
  \end{algorithmic}
  \end{algorithm}

While the original NFSP uses DQN with a replay buffer as its RL algorithm, there can be an on-policy problem. Value-based RL methods including DQN are known to be off-policy algorithms; that is, one can use any data sampled by any behavior policies to train the target policy. However, in NFSP, we cannot use the off-policy data because the opponents' strategies are not stationary. Although we do not need to sample the training data with the target policy, we still need to sample the data along the environment, and the transition rules of the environment now depends on the behavior policy of the opponent. Using a circular replay buffer in self-play requires the strict assumption that the training speed of its opponent is sufficiently slower than the reinforcement learning.

To exploit this fact efficiently, we use a policy gradient algorithm, which is by nature on-policy.
Specifically, we combine NFSP and PPO, a state-of-the-art policy gradient algorithm, applying PPO as the RL method of NFSP. Algorithm~\ref{alg:NFSP_PPO} shows the overview.

In this algorithm, action or training functions are formed into callbacks and registered to a game process. In steps from line~\ref{line:do_step_begin} to line~\ref{line:do_step_end}, multiple games are executed in parallel threads in the process, and one of the registered callback functions is called with appropriate batch information.

In this work, we launch $N$ processes in parallel, and for each process we register agent $p$'s action function that follows the strategy produced by the RL component and other agents' action functions that follow the strategies produced by the SL components, and build multiple game threads in parallel. Here $N$ is the number of agents (in this work $N=2$) and $p$ is the index of a process.

This algorithm is different from the original NFSP, which mixes RL and SL actors and choose either of them at the beginning of each game.
This is again due to an on-policy problem. In the original NFSP, an agent $p$ has four types of experiences, namely, $(\pi_p, \pi_{-p})$, $(\pi_p, \beta_{-p})$, $(\beta_p, \pi_{-p})$, and $(\beta_p, \beta_{-p})$, where $\pi$ is a SL strategy and $\beta$ is a RL strategy. If the RL method is off-policy as in the original NFSP, then we can use all experiences. However, since it is now on-policy, we can only use $(\beta_p, \cdot)$ experiences, which significantly reduces its sample efficiency. Launching $N$ processes in parallel and assigning each agent $i$ for them, we can reduce inefficient data $(\pi_p, \pi_{-p})$ and $(\beta_p, \beta_{-p})$.

Here is another reason for the modification.
Although the ELF platform is general and flexible, there is a difficulty in implementing original NFSP on the platform. In original NFSP, agents need to decide whether they follow the RL component to perform the best response strategy to its opponents, or the SL component to act as the average strategy of its past best response strategies, at the beginning of the game. However, in the ELF platform, in order to calculate the forward computing efficiently, observation data are accumulated, bundled, and sent with a callback function to a corresponding agent as a batch. We thus need to divide the given batch into RL and SL batches, and search for the terminal observation to decide which components to use in each game, spoiling the computing efficiency. The proposed algorithm overcomes the problem and is easier to implement than the original one, because we do not have to decide which component to follow, but just separately build $N$ ELF processes in parallel.

During RL training in line~\ref{line:calc_rl_loss}, $\mathcal{L}_{RL}$ is calculated in almost the same way as in PPO. That is,

\begin{equation}
  \label{eq:rl_cost}
  \mathcal{L}_{RL} = \mathcal{L}_{policy} + \alpha \mathcal{L}_{entropy} + \beta \mathcal{L}_{value}\,,
\end{equation}
where $\mathcal{L}_{policy}$ is the main PPO cost function defined by the negation of the equation \eqref{eq:PPO_loss}, $\mathcal{L}_{entropy} = \sum_{a} \pi_{\theta}(a\middlebar s) \log \pi_{\theta} (a\middlebar s)$ is a bonus term that encourages exploration for the agent, and $\mathcal{L}_{value}$ is a squared mean error between $V_\theta$ and the target value $V_{target} = \hat{A}_t + V_{\theta_{old}}$. The estimator $\hat{A}$ is calculated by $\hat{A}_t = \delta_t + k\delta_{t+1} + \cdots + k^{T-t+1} \delta_{T-1}$, where $\delta_t = r_t + \gamma V_{\theta}(s_{t+1}) - V_{\theta}(s_{t})$.

Following the implementation in the OpenAI Baselines~\citep{OpenAIBaselines2017}, we use the clipped value loss as $\mathcal{L}_{value}$,

\begin{align*}
  V_{clip}(s) =&\; \text{clip}\left( V_{\theta}(s) - V_{\theta_{old}}(s), -\epsilon_{v}, \epsilon_{v} \right) + V_{\theta_{old}}(s)\,,\\
  \mathcal{L}_{value}^{nonclip} =& \left(V_{\theta}(s_{t}) - V_{target}(s_{t})\right)^2 \,,\\
  \mathcal{L}_{value}^{clip} =& \left(V_{clip}(s_t) - V_{target}(s_{t})\right)^2 \,,\\
  \mathcal{L}_{value} =& \max \left(\mathcal{L}_{value}^{nonclip}, \mathcal{L}_{value}^{clip}\right)\,,\\
\end{align*}
and normalize the average and variance of the advantages in a batch.

During SL training in line~\ref{line:calc_sl_loss}, $\mathcal{L}_{SL}$ is calculated by

\begin{equation}
  \label{eq:sl_cost}
  \mathcal{L}_{SL} = -\sum_{a}\pi_{\theta_{RL}}(a|s_t) \log \pi_{\theta}(a | s_t)\,,
\end{equation}
which is the cross entropy between the probability distribution of SL and RL.

For memorizing the experiences in line~\ref{line:memorize_exp}, we use reservoir sampling~\citep{Reservoir1985} as a sampling method for the replay buffer like original NFSP~\citep{NFSP2016}.
A reservoir replay buffer $\mathcal{M}_{RRB}$ maintains $N_{RRB}$ data tuples $\left\{s_{t_i}, \pi_{t_i}\right\}_{i=1,\cdots,N_{RRB}}$ and the number of given tuples $M_{RRB}$. When served $\left\{s_t, \pi_t\right\}$, $\mathcal{M}_{RRB}$ memorizes it with probability $\frac{N_{RRB}}{M_{RRB}+1}$, or otherwise rejects it. When a new tuple is memorized, each old tuple in $\mathcal{M}_{RRB}$ is discarded with equal probability, i.e., in $\frac{1}{N_{RRB}}$. It follows that for any time $T$, each data tuple $\{s_t, \pi_t\}_{t\le T}$ is stored in $\mathcal{M}_{RRB}$ with probability $\frac{N_{RRB}}{M_{RRB}}$, which means that this replay buffer contains a uniform random sample of the given tuples.

We also use the raw self-play method with PPO. The algorithm is the same as the NFSP shown in Algorithm~\ref{alg:NFSP_PPO}, except that the SL\_Actor function and SL training in the Trainer function are omitted and all agents act with the RL\_Actor function.

\section{Experiments}

\subsection{Experimental Settings}

Unless otherwise specified, all experiments are conducted on the following settings.
The batch size is 128 and the batch time is 50; namely, in line~\ref{line:batch_trainer} in Algorithm~\ref{alg:NFSP_PPO} the batch contains $128$ sequences of tuple $\left\{S_\tau, A_\tau, \Pi_\tau, R_\tau\right\}_{\left\{\tau=t,\cdots,t+T-1\right\}}$ where $T=50$, and in line~\ref{line:batch_RL} and \ref{line:batch_SL} the batch contains 128 states. In the reservoir sampling in line~\ref{line:get_batch_from_sl_memory}, we sample 512 states. The frame skip is set to 50, and thus each agent makes its decisions every 50 frames. In a process, 512 games are executed. We use Convolutional Neural Networks (CNNs) as the RL and SL models. Specifically, we use four blocks and some head layers for the CNN, where each block consists of a $3\times 3$ convolutional layer with 64 channels and appropriate zero paddings, batch normalization, and leaky ReLU activation with $\alpha=0.1$. For every two blocks, we use a $2\times 2$ max pooling layer. The head layer is fully-connected and maps the flattened input to an output. There are three heads: $\pi_{SL}$, $\pi_{RL}$, and $V_{RL}$. The heads for $\pi_{SL}$ and $\pi_{RL}$ have nine outputs and a softmax layer to form a probability distribution, whereas the head for $V_{RL}$ has only one output and does not have the softmax layer. The parameters of body blocks for $\pi_{RL}$ and $V_{RL}$ are shared, while $\pi_{SL}$ and $\pi_{RL}$ are not. Note that all agents use the same networks and their $RL$ and $SL$ networks are entirely shared. We use stochastic gradient descent with gradient clipping to optimize the models. The maximum gradient norm is set to 0.5. We use 0.01 and 0.001 for the learning rate of the RL model and the SL model respectively. In the RL loss function in the equation~\eqref{eq:rl_cost}, we use $\alpha=0.01$, $\beta=0.5$, $\gamma=0.99$, $k=0.95$, and $\epsilon_{v}=0.1$.

\subsection{Self-Play and NFSP for Mini-RTS}

We train agents with raw self-play and NFSP, and evaluate them with the win rate against rule-based AIs. In Mini-RTS, there are two rule-based built-in AIs: AI-Simple and AI-Hit-and-Run. AI-Simple simply builds five tanks and then attacks the opponent base. AI-Hit-and-Run is more aggressive and often harasses the opponent with its tanks. A human player has a win rate of 90\% and 50\% against AI-Simple and AI-Hit-and-Run respectively~\citep{elf2017}.

Because the game is a symmetric two-player zero-sum game, if an agent follows a strategy of a Nash equilibrium strategy profile, the agent is never exploitable, and thus it wins at least 50 percent against any strategies.
If the game is sufficiently small, we can evaluate the exploitability of a strategy profile, which is the value that shows how close the strategy profile is to a Nash equilibrium~\citep{Exploit2011}. However, ELF Mini-RTS is too large to calculate it. There are scalable methods to calculate the approximated or bounded exploitability such as local best response teqnique~(LBR)~\citep{LBR2017}. Although LBR can calculate a lower bound of the given strategy profile, it cannot calculate an upper bound. In this work, we simply evaluate the agents with win rates against rule-based AIs, which is an estimator of a lower bound of the exploitability. Each evaluation consists of at least $1000$ games.

\begin{figure}[t]
  \centering
  \begin{subfigure}{\hsize}
    \centering
    \includegraphics[width=75mm]{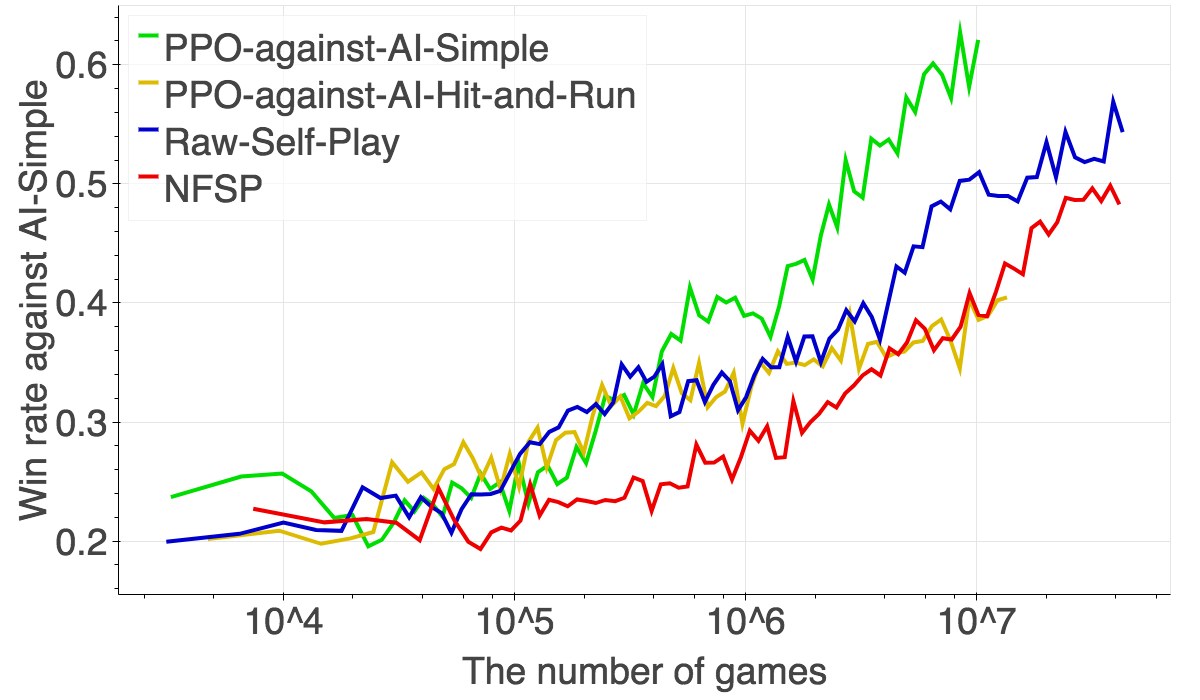}
    \caption{Win rate against AI-Simple.}
    \label{fig:vs_simple}
  \end{subfigure}
  \begin{subfigure}{\hsize}
    \centering
    \includegraphics[width=75mm]{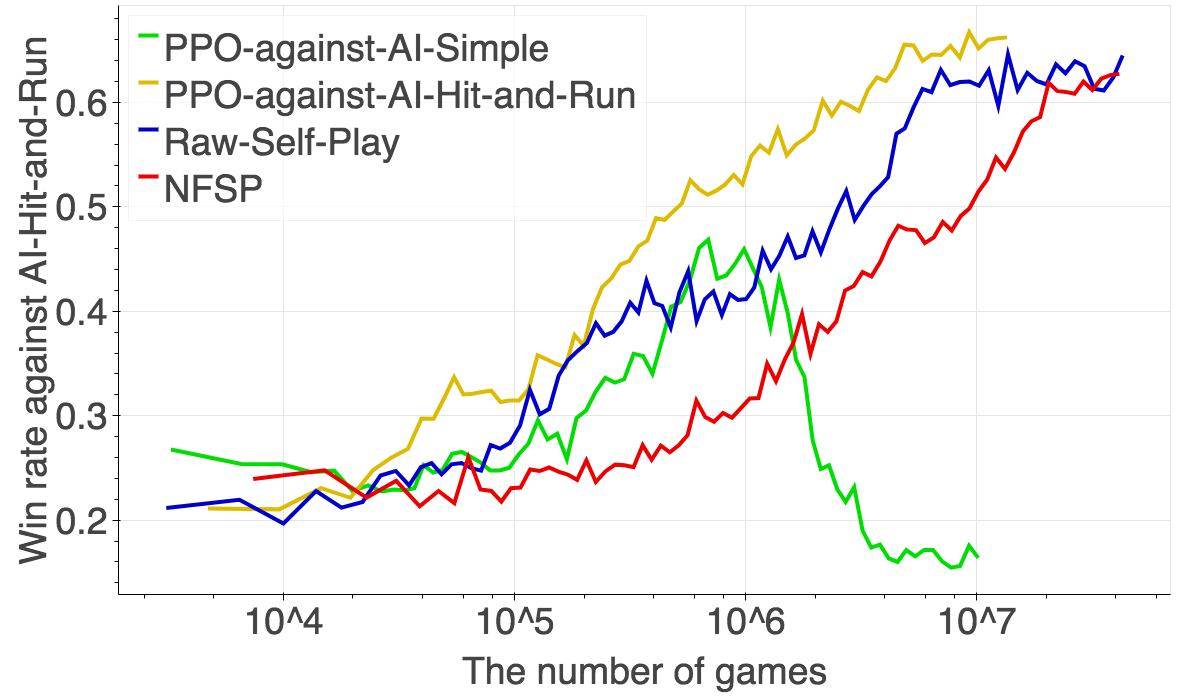}
    \caption{Win rate against AI-Hit-and-Run.}
    \label{fig:vs_har}
  \end{subfigure}
  \caption{Win rate in Mini-RTS with respect to the amount of experience in different methods. The horizontal axis is log-scale. NFSP is shown in the red line.}
  \label{fig:nfsp}
\end{figure}

Figure~\ref{fig:nfsp} shows the results of self-play methods, with the results of PPO agents. Evaluated against AI-Simple, the PPO agent trained against the same AI has the highest win rate. However, it fails to generalize its strategy against AI-Hit-and-Run and thus its strategy is far from Nash equilibria. The PPO agent trained against AI-Hit-and-Run also fails to exploit the AI-Simple.

The win rate of the agent trained with NFSP steadily increases as the number of experienced games increases. Although the rate is lower than the rate of the appropriate PPO agent, the NFSP agent does not fall into a specialized best response strategy but gradually acquires a less exploitable strategy.

The agent trained with raw self-play reaches the same result as the NFSP agent in the AI-Hit-and-Run evaluation, and even better result in the AI-Simple evaluation. Although there is no theoretical guarantee that a self-play algorithm converges, it can reach a Nash equilibrium if it converges~\citep{LearningWithOpponentAwareness2018}, and it is faster than NFSP because NFSP agent has to learn both the best response strategy to its opponent and the averaged strategy.
Note that \citet{RoboschoolBlog2017} show a counter example that a self-play method oscillates and thus does not converge. In this experiment such an oscillation is not observed. 

Note that we do not conduct an experiment with the combined AI, namely, an AI that acts as AI-Simple in 50\% and acts as AI-Hit-and-Run in 50\%, unlike \citet{elf2017}, because we evaluate the strategies with these built-in AIs and we have to make at least one of them unknown to the trained agent to evaluate its performance against unseen opponents.

From Figure~\ref{fig:nfsp} we can see that even the highly specialized agent wins in at most 65\% of the games. This is because the Mini-RTS game has considerable randomness at the beginning of the game. When the game starts, resources, bases, and units are randomly placed in the game field. Because of the frame skip, if an agent has no tank and its opponent has some tanks at the beginning of the game, and the opponent decides to attack with them, the agent has no way to defend against the rush. Even the agent trained with PPO in $10^7$ games against a pure random agent loses in 29\% of the game against the same random agent.

\subsection{Analysis of the acquired agents}

We further analyze the acquired agents. To evaluate how exploitable the agent is, we train another PPO agent against the target agent. If the PPO algorithm converges to its optimal strategy, the win rate of the agent is equal to the exploitability of the target agent in imperfect recall settings.

The results are shown in Table~\ref{table:ppo_against_eachAI}. Compared with other agents, the self-play agents are less exploitable, and do not lose over 50 percent against the PPO algorithm. This result suggests that the obtained agents are not exploitable by strategies that do not use the past histories of observations.

\begin{table}[t]
  \centering
  \begin{tabular}{|l||c|} \hline
    Agent & Win rate \\ \hline \hline
    Random & $0.71$ \\ \hline
    AI-Simple & $0.62$ \\ \hline
    AI-Hit-and-Run & $0.65$ \\ \hline
    PPO against AI-Simple & $0.80^\dagger$ \\ \hline
    PPO against AI-Hit-and-Run & $0.56$ \\ \hline
    Raw self-play with PPO & $\bm{0.45}$ \\ \hline
    NFSP with PPO & $\bm{0.44}$ \\ \hline
  \end{tabular}
  \caption{Win rates of the trained PPO agent against each AI. All agents are trained with $10^7$ games except the results with $\dagger$, which means the number of training games is less than $10^7$. The self-play agents have the lowest win rate, and hence they are less exploitable.}
  \label{table:ppo_against_eachAI}
\end{table}

\begin{figure}[t]
  \centering
  \begin{subfigure}{0.47\hsize}
    \centering
    \includegraphics[width=39mm]{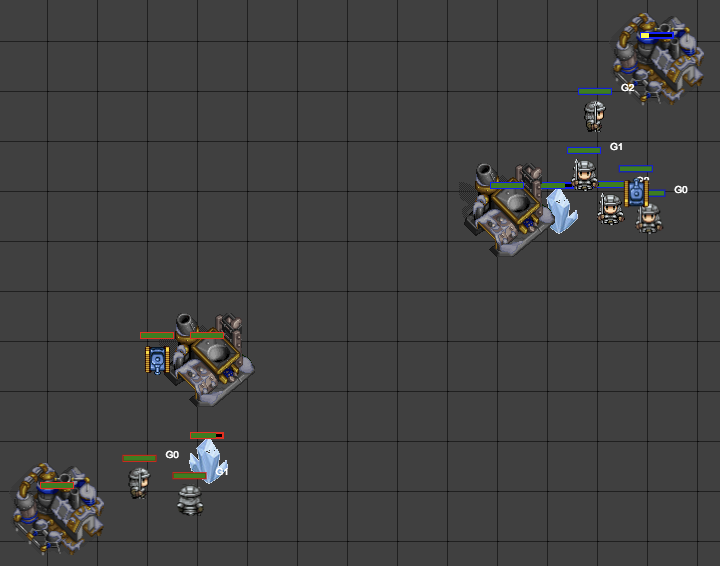}
    \caption{}
    \label{fig:NFSP_game_1_1}
  \end{subfigure}
  \hspace{0.02\hsize}
  \begin{subfigure}{0.47\hsize}
    \centering
    \includegraphics[width=38mm]{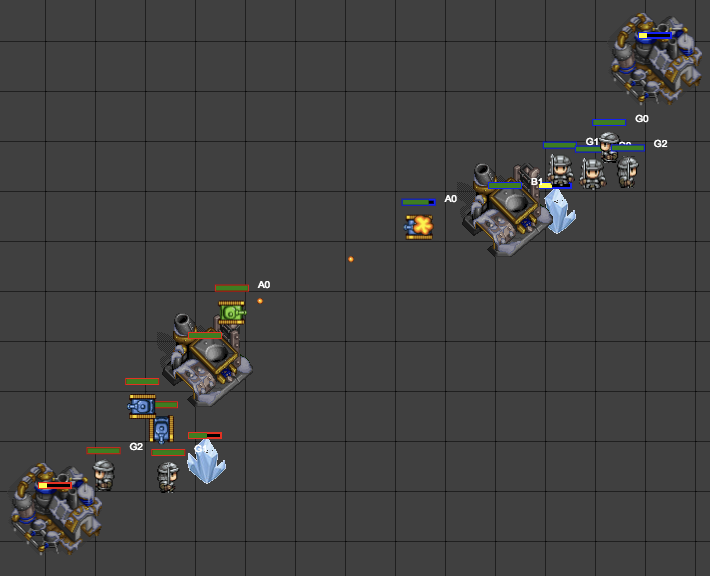}
    \caption{}
    \label{fig:NFSP_game_1_2}
  \end{subfigure}
  \caption{Screenshots of a game between the NFSP agent (red, bottom left) and the PPO agent trained against it (blue, top right). The blue tanks are melee attackers and the green tanks are range attackers. The NFSP agent (a) first builds a melee attacker, which is suited for defense, then (b) builds a range attacker, which is suited for attacking, and rushes to the opponent's base.}
  \label{fig:NFSP_game_1}
\end{figure}

We observe the details of some games between the NFSP agent and the PPO agent trained against the NFSP agent. Figure~\ref{fig:NFSP_game_1} shows some screenshots of the game. The NFSP agent first builds two melee attackers, next builds a range attacker, and then rushes to the opponent's base. Because of the fog-of-war, agents cannot be aware of its opponent's attack until the opponent's tanks get closer, and thus melee attackers are suited for defense while range attackers are suited for attacking. Hence, the behavior of the NFSP agent is very rational for humans: first build some defense units to prepare for its opponent's attacking, secondly build an attacking unit with keeping the previously built defense units, and finally attack the opponent's base with all tanks.

\begin{figure}[t]
  \centering
  \begin{subfigure}{0.47\hsize}
    \centering
    \includegraphics[width=39mm]{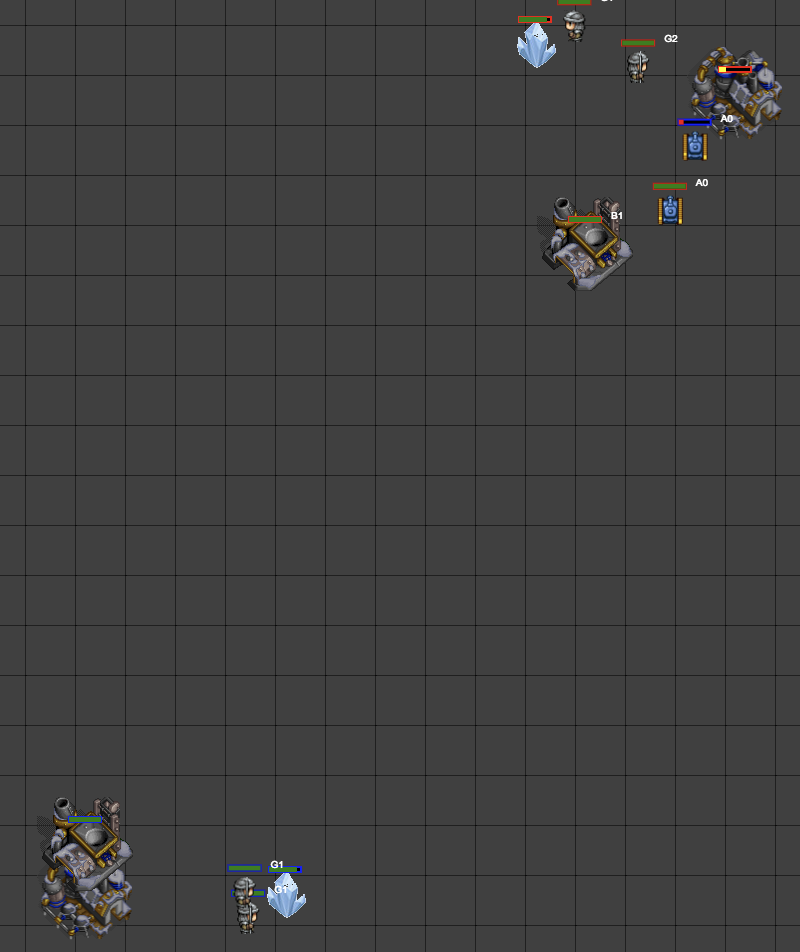}
    \caption{}
    \label{fig:NFSP_game_2_1}
  \end{subfigure}
  \hspace{0.02\hsize}
  \begin{subfigure}{0.47\hsize}
    \centering
    \includegraphics[width=39mm]{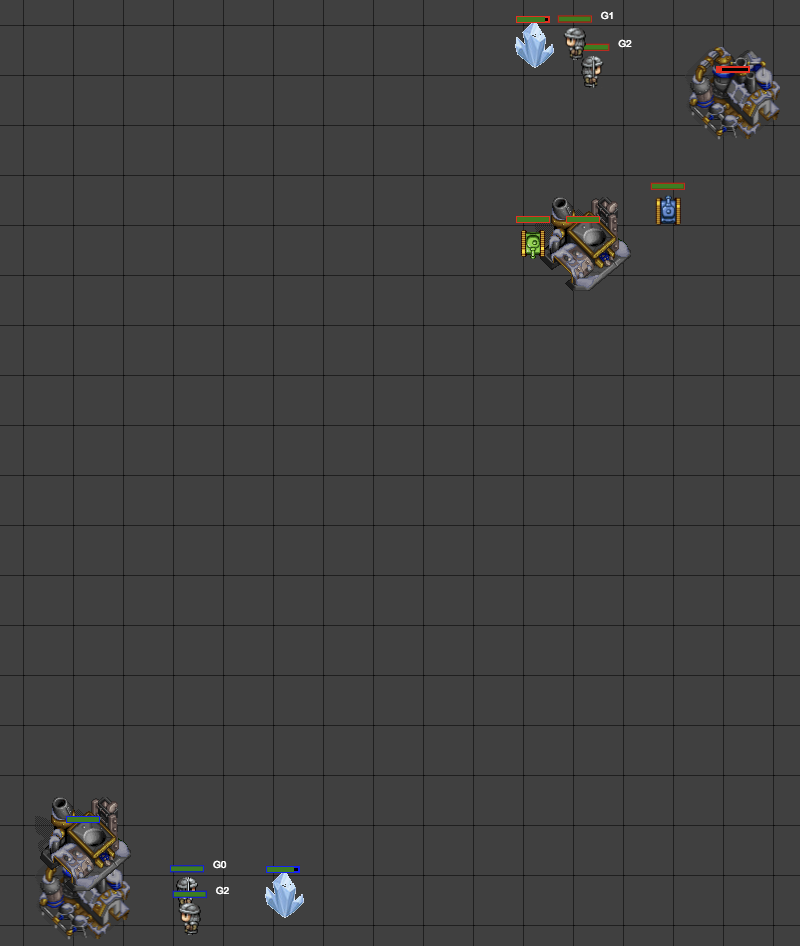}
    \caption{}
    \label{fig:NFSP_game_2_2}
  \end{subfigure}
  \caption{Screenshots of a game between the NFSP agent (red, top right) and the PPO agent trained against it (blue, bottom left). (a) The NFSP agent's base is now very fragile because of its opponent's attack. (b) It builds a range attacker (to attack) and not a melee attacker (to defend), because in this settings agents cannot attack with a part of its tanks but must attack with all tanks, and thus its opponent must have now no tanks.}
  \label{fig:NFSP_game_2}
\end{figure}

We show another example. In the game shown in Figure~\ref{fig:NFSP_game_2}, at the initial state the NFSP agent has a large disadvantage due to the randomness of ELF games: it does not have a barrack while its opponent does, and it does not know the disadvantage because of the fog-of-war. Having the disadvantage, the NFSP agent is attacked by the opponent's tanks, but it builds a range attacker (suited for attacking) unit, and successfully counterattacks with it. Because in this game an agent must attack with all tanks it has, the NFSP agent knows that the opponent has now no tanks. Although an agent does not know or memorize the state of the opponent, the NFSP agent successfully exploits the rule and estimates the unknown state without any prior knowledge or even any built-in rule-base AIs.

\subsection{Pretraining NFSP with Raw Self-Play}

In the previous experiments, we observe that the NFSP agent successfully acquires a less exploitable strategy profile, but the learning process is slower than other methods. In contrast, the raw self-play algorithm is fast but lacks the guarantee of convergence. If the NFSP agent can be pretrained with the raw self-play algorithm, we can take the advantages of both algorithms.

This insight is also seen in CounterFactual Regret Minimization+ (CFR+)~\citep{CFR+2014}. In a regret matching algorithm, which is a basis of CFR+, the average strategy of the regret-based strategies converges to a Nash equilibrium. This schema is similar to the fictitious play: in a fictitious play algorithm, we compute a best response strategy instead of the regret-based strategy, and average them. In CFR+, \citet{CFR+2014} uses delayed averaging, namely, accumulates the strategies from the middle of them. It significantly improves the result. The pretraining of NFSP is regarded as a kind of delayed averaging, because in PPO we do not accumulate the strategies and then switch to NFSP and begin to averaging them.

\begin{figure}[t]
  \centering
  \begin{subfigure}{\hsize}
    \centering
    \includegraphics[width=75mm]{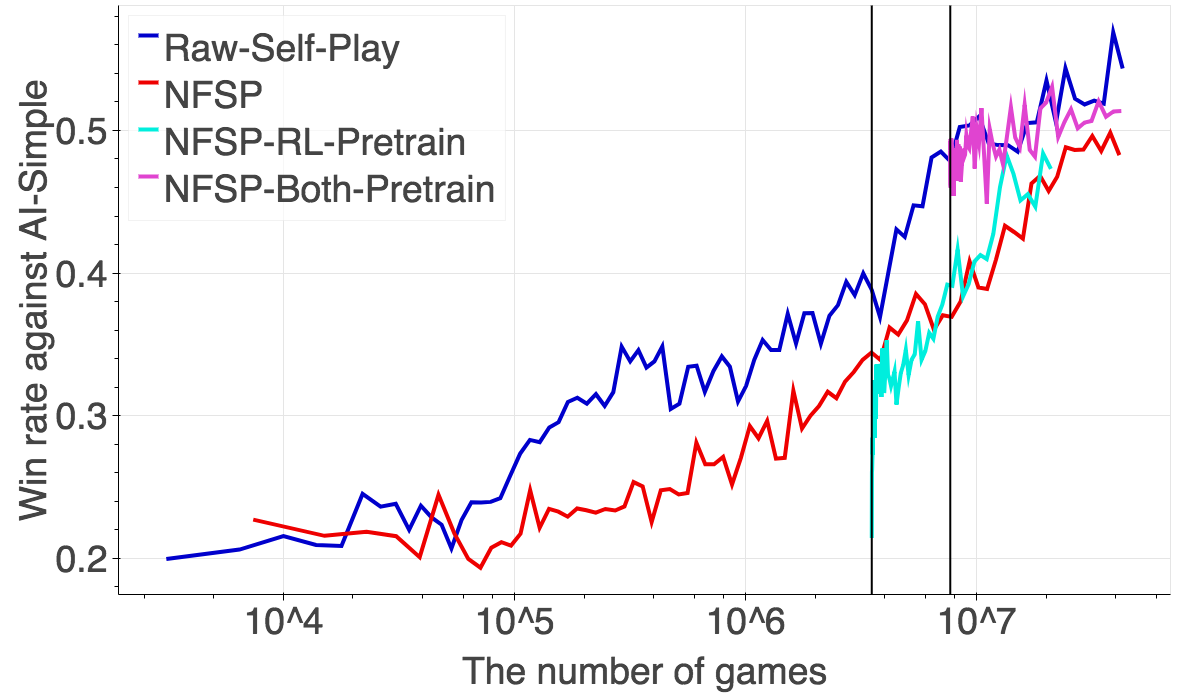}
    \caption{Win rate against AI-Simple.}
    \label{fig:pretrain_vs_simple}
  \end{subfigure}
  \begin{subfigure}{\hsize}
    \centering
    \includegraphics[width=75mm]{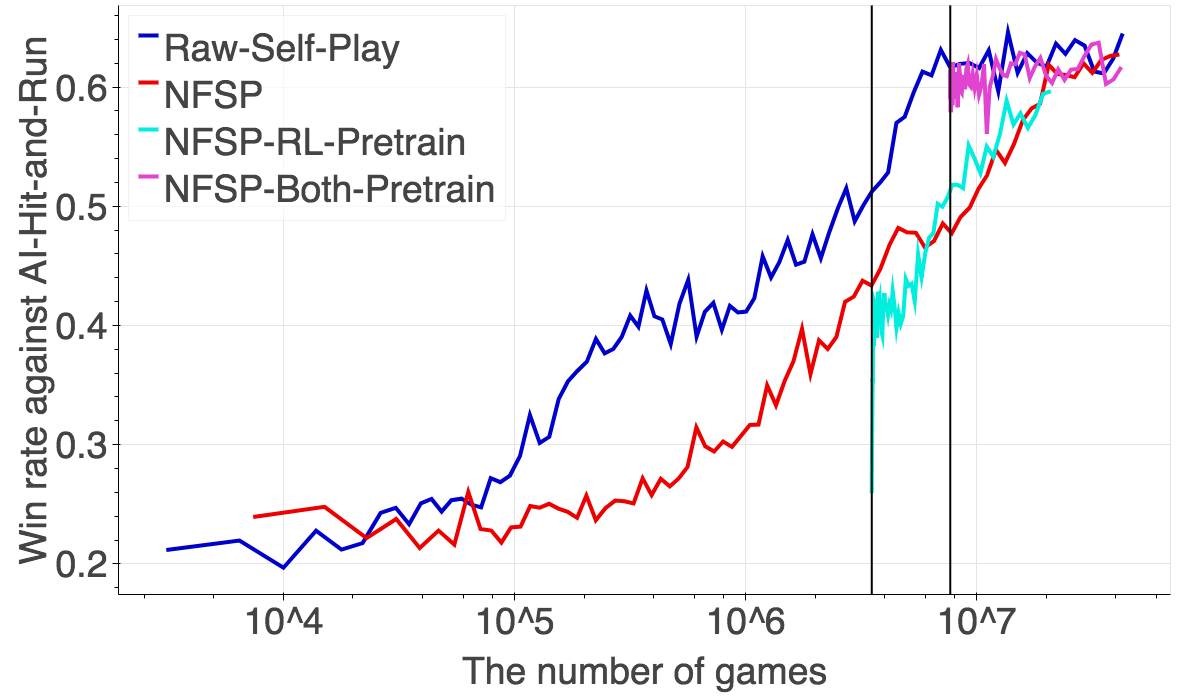}
    \caption{Win rate against AI-Hit-and-Run.}
    \label{fig:pretrain_vs_har}
  \end{subfigure}
  \caption{Win rate of the pretrained agents with respect to the amount of experience. The horizontal axis is log-scale. The blue line shows the base self-play results, the light-blue line is the NFSP whose RL component is pretrained, the purple line is the NFSP whose both RL and SL components are pretrained, and the red line is the non-pretrained NFSP. The vertical black lines show the beginning of the pretraining.}
  \label{fig:nfsp_pretrain}
\end{figure}

First we pretrain the RL model in NFSP with raw self-play. The light-blue lines in Figure~\ref{fig:nfsp_pretrain} show the result. Although the learning is slightly faster than the NFSP agent, it does not improve the performance as we expected. This result can be explained as follows: in NFSP, the SL component averages the RL strategy and the RL component computes the best response strategy to the SL strategy. When we pretrain the RL strategy, the SL component can accumulate the pretrained strategies, and thus its learning process is accelerated. However, because the RL component computes the best response to the non-pretrained SL component, training in the RL component is not accelerated at all, making it fail to improve the performance.

To solve this problem, we also pretrain the SL model with the same parameters used in the RL model. We use the $\pi_{RL}$ head of the PPO agent to pretrain the $\pi_{SL}$ head of the NFSP agent, and simply discard the $V_{RL}$ head of the PPO agent.
The purple lines in Figure~\ref{fig:nfsp_pretrain} show the result. Although the result is worse than the raw self-play, it successfully maintains the result of its base strategy, and is even slightly fine-tuned from the strategy in the AI-Simple evaluation. This result suggests that we can extend the results from a faster but more unstable self-play algorithm as pretraining for NFSP.

\section{Conclusion and Future Work}
In this paper, we regard the Mini-RTS game as a two-player zero-sum extensive-form game, and apply self-play methods. The obtained agent is less exploitable for the PPO algorithm than other best response-based agents. We also observe that the obtained agent performs rationally to humans.

The contribution of this paper is that we show that NFSP can be combined with policy gradient reinforcement learning and be applied to Mini-RTS, which can be a first step toward solving more realistic and complex RTS games.
We also show that we can improve the scalability of NFSP by pretraining the models with simple self-play using policy gradients, which is faster but lacks the theoretical guarantee of convergence. It significantly reduces the computational time and could be applied even when the self-play algorithm oscillates.
However, the experimental results show that the learning process of NFSP is much slower than raw self-play with PPO, and actually raw self-play successfully acquires reasonable strategies despite its lack of convergence guarantees. We will further analyze the results and the differences between NFSP and raw self-play methods.

In this paper we do not have the agents memorize past histories. This makes the game essentially imperfect recall, which breaks the assumption of the FSP. To solve this, we could use a recurrent neural networks as a controller of the RL component as in~\citet{DRQN2015}. However, to ensure that the game is a perfect recall game, we need to use the same memorizing architecture for the SL reservoir replay buffer, which significantly reduces the size of the buffer. We will also further investigate to solve this problem as future work.

\bibliography{all}
\bibliographystyle{aaai}
\end{document}